\documentclass[runningheads]{llncs}
\usepackage{graphicx}
\usepackage{cite}
\usepackage{amsmath,amssymb,amsfonts}
\usepackage{algorithmic}
\usepackage[caption=false, font=footnotesize]{subfig}
\usepackage{lipsum}
\usepackage{textcomp}
\usepackage{verbatim}
\usepackage{booktabs,caption}
\usepackage{multirow}
\usepackage{makecell}
\usepackage{bm}
\usepackage{color}
\usepackage[table]{xcolor}
\usepackage[ruled,vlined]{algorithm2e}
\usepackage[misc]{ifsym}

\newcommand{\etal}{\textit{et al}.}
\definecolor{anti-flashwhite}{rgb}{0.95, 0.95, 0.96}
\definecolor{blueish}{rgb}{0.0, 0.3, .6}
\definecolor{light_cyan}{rgb}{0.88,1,1}
\begin{document}
%
% -----------------------------------------------------
\title{Iterative Online Image Synthesis via Diffusion Model for Imbalanced Classification}
% -----------------------------------------------------
\author{Shuhan Li$^{\dag}$\textsuperscript{\Letter}, Yi Lin$^{\dag}$, Hao Chen, and Kwang-Ting Cheng}
% ----------------------------------------
\institute{The Hong Kong University of Science and Technology, Hong Kong, China \\
\email{slidm@connect.ust.hk}
}

% \titlerunning{IOIS}
\titlerunning{Iterative Online Image Synthesis for Imbalanced Classification}
\authorrunning{S. Li and Y. Lin, \etal}
\maketitle              % typeset the header of the contribution
\def\thefootnote{$\dag$}\footnotetext{Equal contribution; \Letter~corresponding author.}

\begin{abstract}
Accurate and robust classification of diseases is important for proper diagnosis and treatment. However, medical datasets often face challenges related to limited sample sizes and inherent imbalanced distributions, due to difficulties in data collection and variations in disease prevalence across different types. In this paper, we introduce an Iterative Online Image Synthesis (IOIS) framework to address the class imbalance problem in medical image classification. 
Our framework incorporates two key modules, namely Online Image Synthesis (OIS) and Accuracy Adaptive Sampling (AAS), which collectively target the imbalance classification issue at both the instance level and the class level. 
The OIS module alleviates the data insufficiency problem by generating representative samples tailored for online training of the classifier. 
On the other hand, the AAS module dynamically balances the synthesized samples among various classes, targeting those with low training accuracy. 
To evaluate the effectiveness of our proposed method in addressing imbalanced classification, we conduct experiments on the HAM10000 and APTOS datasets. The results obtained demonstrate the superiority of our approach over state-of-the-art methods as well as the effectiveness of each component.
The source code will be released upon acceptance.
\keywords{Image synthesis \and Imbalanced classification \and Diffusion model.}
\end{abstract}

\section{Introduction}
Image classification is an essential task for medical image analysis and has wide applications in medical datasets, such as distinguishing benign or malignant tumors~\cite{cui2022fmrnet,saleh2020brain}, grading specific diseases~\cite{wang2021detection,yan2022nuclei}, and diagnosing various diseases~\cite{li2023dynamic,baccouche2020ensemble}. 
However, medical datasets often encounter challenges related to insufficient sample sizes and inherent imbalanced distributions, which can be attributed to difficulties in data collection and variations in disease prevalence across different types~\cite{banik2021mitigating}. 
These challenges can lead to poor generalization and biased predictions, which can be detrimental to the performance of deep learning models.

To address the challenges of class imbalance, existing solutions can be categorized into three main groups: re-weighting, re-sampling, and data synthesis. Re-weighting methods~\cite{lin2017focal,cui2019class,tan2020equalization,tan2021equalization} aim to balance the loss between the majority and minority classes. 
An example is Focal Loss~\cite{lin2017focal}, which adjusts hyper-parameters to increase the loss for minority classes while decreasing it for majority classes. This approach encourages the model to focus more on challenging samples. 
On the other hand, re-sampling and data synthesis methods directly modify the distributions of the original dataset. Re-sampling methods~\cite{mahajan2018exploring,kang2019decoupling} involve oversampling or undersampling techniques. Oversampling increases the sampling probability of minority classes, whereas undersampling decreases the probability of majority classes. Data synthesis methods~\cite{shen2023cellgan,carrasco2022assessing,yang2021ida} utilize generative models such as generative adversarial networks (GANs) to generate new images for minority classes. These generated samples help balance the sample size across classes, thereby addressing the class imbalance issue.

Recently, the diffusion model (DM) has emerged as a potent generative model that exhibits superior performance compared to GANs~\cite{dhariwal2021diffusion,rombach2022high}.
Building on the success of DM, various methods have demonstrated the potential of DM in improving medical image classification by synthesizing training samples~\cite{zhong2023meddiffusion,ye2023synthetic}, which is particularly useful for medical image classification tasks with limited training data.
However, existing methods typically generate images independently before commencing the training process, keeping the training images unchanged throughout.
This approach can lead to overfitting on the synthetic images due to the discrepancy between the generator and the downstream classifier.
Moreover, the portion of synthetic images for each class is often determined manually in these methods, which may not align with the dynamic requirements of the classifier for each class during the training process.

To address aforementioned issues, in this paper, we propose a novel \textbf{I}terative \textbf{O}nline \textbf{I}mage \textbf{S}ynthesis framework, named \textbf{IOIS}, to address the class imbalance problem in medical image classification. The main contributions of our work are summarized as follows: 1) At the instance level, we introduce an Online Image Synthesis (OIS) module to alleviate the data insufficient problem.
For each epoch, we employ the gradient of the training classifier as guidance for the diffusion model to generate samples tailored for online training. As the classifier develops during training, the synthesized images become more representative of their respective classes, which in turn benefits the online training process iteratively.
2) At the class level, we propose an Accuracy Adaptive Sampling (AAS) module to dynamically balance the synthesized samples among various classes.
AAS aims to generate more samples for classes with low training accuracy, addressing the class imbalance issue in a targeted manner. 3) We conduct experiments on two public medical image datasets to evaluate the effectiveness of our method. The results demonstrate the superiority of our approach compared to state-of-the-art methods, as well as the effectiveness of each component. 

% ----------------------------------------
\begin{figure*}[t]
\centering
\includegraphics[width=.98\textwidth]{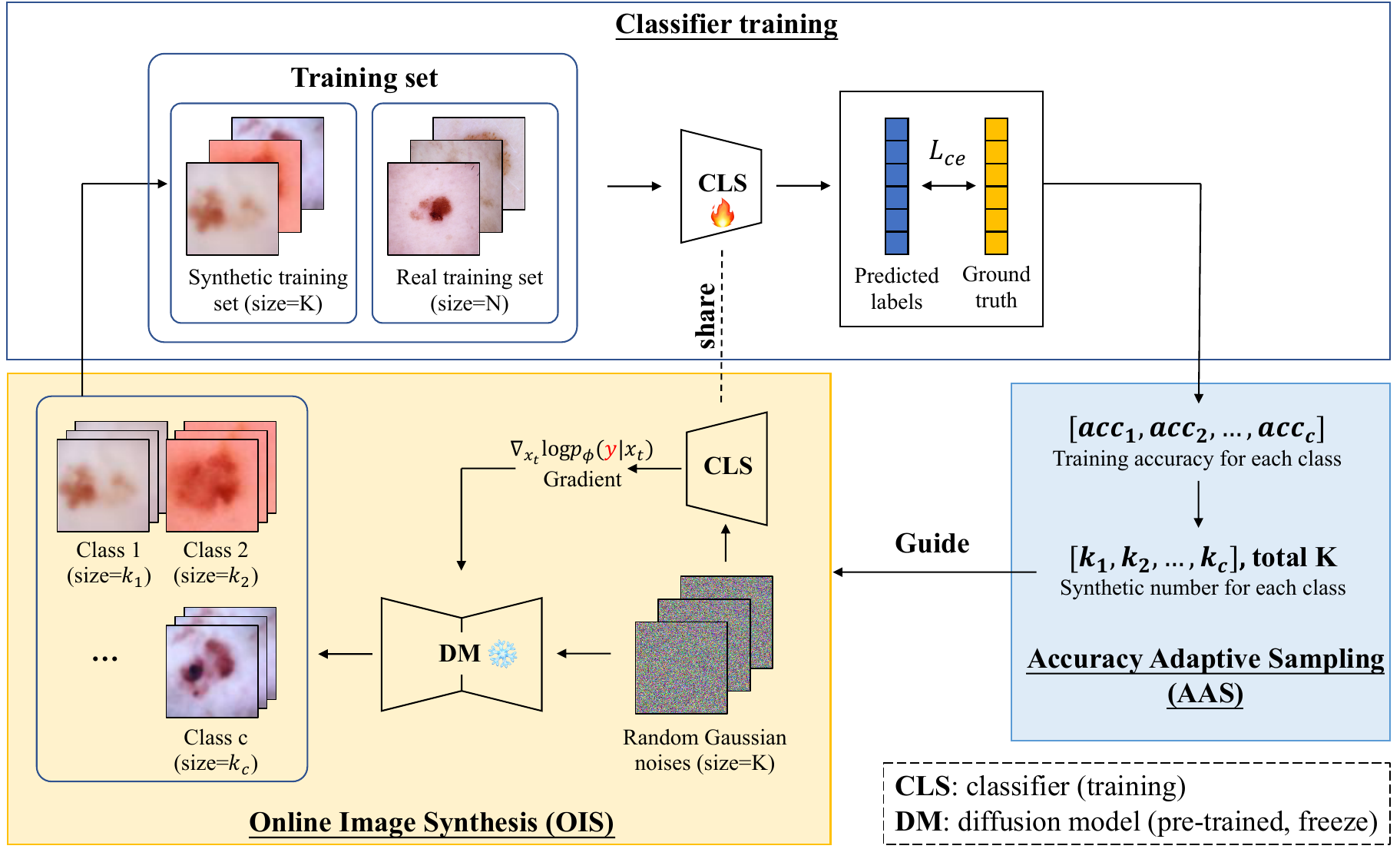}
\caption{
The framework of the proposed IOIS. It comprises three conponents, which are iteratively performed: training classifier, determining class distribution of synthetic images via AAS, and synthesizing images via OIS.
}
\label{fig1}
\end{figure*}
% ----------------------------------------

\section{Methodology}
As depicted in Figure~\ref{fig1}, the proposed Iterative Online Image Synthesis (IOIS) framework comprises three key components: classifier, the Online Image Synthesis (OIS) module, and the Accuracy Adaptive Sampling (AAS) module. 
First, a conventional classifier training process is conducted, where we adopt ResNet-50 as the backbone network.
Then, the OIS module is introduced to generate augmented data tailored to the classifier during the training process.
Additionally, the AAS module is proposed to update the distribution of synthesized images in the diffusion model based on the feedback from the classifier's performance.
In the following, we will detail each component.

\subsection{Diffusion Model}
\label{method1}
Diffusion model (DM)~\cite{ho2020denoising} is a probabilistic generative model that aims to learn the probability distribution of a given dataset. It defines the \textit{forward} process to gradually add Gaussian noise to a clean image and the \textit{reverse} process to recover the noised image step by step. 

In the \textit{forward} process, we gradually add Gaussian noise to an image sampled from a data distribution $\mathbf{x}_{0}\sim q(\textbf{x})$ in $T$ steps to obtain noisy samples $\{\mathbf{x}_1,...,\mathbf{x}_t,...,\mathbf{x}_T\}$, where $t\in \{1,..., T\}$. We use $q(\mathbf{x}_t|\mathbf{x}_{t-1})$ to represent forward process at $t$-th step,
\begin{equation}
\begin{aligned}
q(\mathbf{x}_t|\mathbf{x}_{t-1})=\mathcal{N}(\mathbf{x}_t;\sqrt{1-\beta_t}\mathbf{x}_{t-1},\beta_t\mathbf{I}),
\label{forward_one_step}
\end{aligned}
\end{equation}
where $\beta_t\in (0,1)$ denotes the variance which controls the step size for $t$-th step, and $\mathbf{I}$ is the identity matrix. 
\begin{comment}
\begin{equation}
\begin{aligned}
\mathbf{x}_t=\sqrt{\overline{\alpha}_t}\mathbf{x}_{0}+\sqrt{(1-\overline{\alpha}_t)}\bm{\epsilon},
\label{forward_t_steps}
\end{aligned}
\end{equation}
where $\alpha_t=1-\beta_t$, $\overline{\alpha}_t=\prod^{t}_{i=1}\alpha_i$, and $\bm{\epsilon}\sim\mathcal{N}(0,\textbf{I})$.
\end{comment}
In the \textit{reverse} process, we train a model $\bm{\epsilon}_{\theta}$ to predict the noise added on $\mathbf{x}_t$. 
Following~\cite{ho2020denoising}, the simplified training objective is to minimize the mean square error loss:
\begin{equation}
\begin{aligned}
L_{diff}=E_{t,\mathbf{x}_0,\bm{\epsilon}}[\|\bm{\epsilon}-\bm{\epsilon}_{\theta}(\mathbf{x}_t,t)\|^2].
\label{mse_loss}
\end{aligned}
\end{equation}
In the inference stage, we start from a randomly sampled pure Gaussian noise $\mathbf{x}_T$ and remove one step noise by
\begin{equation}
\begin{aligned}
\mathbf{x}_{t-1}=\frac{1}{\sqrt{\alpha_t}}\left(\mathbf{x}_t-\frac{1-\alpha_t}{\sqrt{1-\overline{\alpha}_t}}\bm{\epsilon}_{\theta}(\mathbf{x}_t,t)\right)+\sigma_t\mathbf{z}, \mathbf{z}\sim\mathcal{N}(0,\mathbf{I}).
\label{predict_xt}
\end{aligned}
\end{equation}
After $T$ steps calculation, we can finally obtain $\mathbf{x}_0$ from $\mathbf{x}_T$.

This label-free pre-training approach presents promising prospects, as it allows us to harness the vast reserves of unlabeled data and facilitates the utilization of a wide array of pre-trained diffusion models, thereby enhancing the scalability of our approach.
Once the diffusion model is pre-trained, we freeze the parameters of the DM and only perform inference to generate augmented data for the classifier training.
The class distribution of the synthetic images is dynamically determined by the proposed OIS and AAS modules, which we will describe in the following sections.

\subsection{Online Image Synthesis with Classifier Guidance}
\label{method2}
In contrast to existing methods that solely rely on a pre-trained diffusion model to generate augmented data for training the classifier, we propose a novel module called online image synthesis (OIS). OIS addresses the adaptivity limitations in the classifier by producing augmented data specifically tailored to the classifier during the training process.
The key idea of the OIS module is to utilize the development of the classifier to synthesize more representative images for each class. 
Specifically, during the classifier training process, instead of solely using the real images to guide the DM inference as in Eq.~(\ref{predict_xt}), we further incorporate the classifier's feedback by using the classifier's gradient of $\mathbf{x}_t$ from the output of diffusion model:
\begin{equation}
\begin{aligned}
\hat{\bm{\epsilon}} = \bm{\epsilon}_{\theta}(\mathbf{x}_t,t) - s \cdot \nabla_{\mathbf{x}_t}\log p_{\phi}(y|\mathbf{x}_t).
\label{cls_guidance}
\end{aligned}
\end{equation}
Then, $\mathbf{x}_{t-1}$ is derived by replace the $\bm{\epsilon}_{\theta}$ with the updated value in Equation~\ref{predict_xt}. With the iteration of several steps, the intermediate image $\mathbf{x}_{t}$ becomes close to the distribution of class $y$ as the decreasing of the noise level. In the end, the recovered image $\mathbf{x}_{0}$ belongs to class $y$.

\subsection{Accuracy Adaptive Sampling for Imbalanced Data}
\label{method3}
To achieve a better balance of samples among different classes, we propose an Accuracy Adaptive Sampling (AAS) module designed to dynamically determine the class distributions of synthetic images based on the accuracy of each class.
Specifically, for each epoch of classifier training, we compute the training accuracy for each class, resulting in a vector $[\text{acc}_1, \text{acc}_2, ..., \text{acc}_c]$, where $c$ represents the number of classes. Based on the training accuracy, we calculate the synthetic number, denoted as $k_i$, for each class $i$. The value of $k_i$ is determined by the following equation:
\begin{equation}
\begin{aligned}
[k_1, k_2, ..., k_c] = \text{Softmax}([1-\text{acc}_1, 1-\text{acc}_2, ..., 1-\text{acc}_c])*K,
\label{number_i}
\end{aligned}
\end{equation}
where $K$ denotes the total size of synthetic images for all the classes. The obtained number $k_i$ for class $i$ reflects the performance, that is the lower accuracy of the class, the more samples will be generated in the third step. As a result, the classes with lower accuracy obtain more training samples in the next epoch, which can improve the following performance.
Algorithm~\ref{alg:algorithm1} outlines the whole pipeline of our proposed approach. 

\begin{algorithm}[t]
	\caption{Framework of IOIS, given a diffusion model $\bm{\epsilon}_{\theta}$, and gradient scale $s$.}
	\label{alg:algorithm1}
	\KwIn{Real training images $\mathbf{x}_{re}\sim D_{real}$, random Gaussian noise $\mathbf{x}_{T}\sim \mathcal{N}(0,\mathbf{I})$}
	\KwOut{Labels of images in the test set}  
	\BlankLine
	Initialize the parameters of classifier $\bm{\phi}$ with pre-trained ImageNet weights;
 
	\For{\textnormal{each epoch}}{
        Update $\bm{\phi}$;
        
        Compute accuracy for each class on the training set [$acc_1$,...,$acc_c$] ;
        
        Compute the synthetic numbers for each class by Equation~\ref{number_i} [$k_1,...,k_c$];

        Sample $K$ images with diffusion model $\bm{\epsilon}_{\theta}$ and the classifier $\bm{\phi}$;

        Merge $K$ synthesized images and the real training images as the training set for the next epoch.
        
        }

\end{algorithm}

\vspace{-1mm}	

\section{Experiments}
\subsection{Datasets and Evaluation Metrics}
We evaluate our method on two public medical image datasets with imbalanced class distributions. The \textbf{HAM10000} dataset~\cite{tschandl2018ham10000} consists of 10,015 dermatoscopic images from 7 skin lesion categories. We follow the data splitting settings in~\cite{li2023learning}, where the proportions of the training set, the validation set, and the test set are 70\%, 10\%, and 20\%, respectively. The \textbf{APTOS} dataset~\cite{aptos2019-blindness-detection} includes 3,662 retinal fundus images which are divided into five categories. 
Similarly, we split 70\%, 10\%, and 20\% of the dataset as training, validation, and test set. The imbalanced ratios~\cite{li2022flat} of HAM10000 and APTOS are 59 and 9, respectively. We leverage three evaluation metrics for imbalanced classification which are Macro-F1, Balanced Accuracy (B-ACC), and Matthew's correlation coefficient (MCC).

\subsection{Impelementation Details}
We utilize ResNet-50 as the backbone architecture and initialize the parameters with the pre-trained model on the ImageNet dataset. To train the model, we set the batch size to 32, and resize the input images to 224 $\times$ 224. We apply the standard augmentations including random cropping, random rotation (-30 degrees to +30 degrees), and horizontal flipping. We choose the SGD optimizer with a 0.0125 initial learning rate for 200 epochs of training. We decay the learning rate by 0.1 times at 60, 120, and 180 epochs. 
To prevent overfitting, we select the test model based on the maximum Macro-F1 on the validation set.

\subsection{Comparisons with State-of-the-art Methods}

\begin{table*} [t]
\centering
\renewcommand\arraystretch{1.05}
\caption{Comparisons with state-of-the-art methods on HAM10000 and APTOS datasets. 
Results of other methods are re-implemented under the same setting.}
\setlength{\tabcolsep}{5pt}{
\begin{tabular}{r|ccc|ccc}

\toprule
\multirow{2}{*}{Method} & \multicolumn{3}{c|}{HAM10000 Dataset} & \multicolumn{3}{c}{APTOS Dataset} \\ 
& Macro-F1 & B-ACC & MCC & Macro-F1 & B-ACC & MCC \\
\midrule
CE Loss & 77.78 & 76.78 & 76.11 & 69.05 & 67.91 & 76.79 \\
Focal Loss~\cite{lin2017focal} & 77.72 &  77.54 & 75.04 & 70.22 & 67.82 & 77.47 \\
CB-Focal~\cite{cui2019class} & 77.96 & 79.90 & 75.31 & 70.54 & 70.36 & 77.19 \\
Sqrt-RS~\cite{mahajan2018exploring} & 78.55 & 79.44 & 76.13 & 70.02 & 68.61 & 77.68 \\
PG-RS~\cite{kang2019decoupling} & 79.09 & 78.45 & 77.78 & 70.77 & 69.77 & 76.98 \\
Cell-GAN~\cite{shen2023cellgan} & 78.89 & 79.43 & 77.74 & 71.10 & 70.26 & 78.86 \\
StyleGAN2-ADA~\cite{carrasco2022assessing} & 80.03 & 80.14 & 78.82 & 72.48 & 71.05 & 79.26 \\
\midrule 
\rowcolor{light_cyan}
Ours (+offline) & 80.09 & 79.97 & 78.79 & 72.79 & 70.86 & 79.95 \\
\rowcolor{light_cyan}
Ours (+OIS) & 80.89 &  80.20 & 79.03 & 72.72 & 71.08 & 80.17 \\
\rowcolor{light_cyan}
Ours (+OIS, AAS) & \textbf{81.97} & \textbf{81.50} & \textbf{80.64} & \textbf{73.02} & \textbf{72.27} & \textbf{80.76} \\
\bottomrule
\end{tabular}
}
\vspace{-1mm}
\label{table1}
\end{table*}

We conducted a comprehensive comparison of our proposed method with several state-of-the-art approaches on the HAM10000 and APTOS datasets. To ensure a fair evaluation, we re-implemented the competing methods using the same backbone architecture and data augmentation strategies.
The comparison results are summarized in Table~\ref{table1}. The compared methods were specifically designed to address the issue of imbalanced classification, including re-weighting techniques (Focal Loss~\cite{lin2017focal} and CB-Focal~\cite{cui2019class}), re-sampling approaches (Sqrt-RS~\cite{mahajan2018exploring} and PG-RS~\cite{kang2019decoupling}), and GAN-based synthetic methods (Cell-GAN~\cite{shen2023cellgan} and StyleGAN2-ADA~\cite{carrasco2022assessing}).
Compared to the baseline method (CE Loss), which employs cross-entropy loss for training, other methods demonstrate improvements across all three metrics in general.
However, the re-weighting and re-sampling techniques achieve these enhancements for minority classes at the expense of accuracy for majority classes. For instance, on the HAM10000 dataset, the CB-Focal method obtains a 3.12\% improvement in B-ACC while experiencing a 0.80\% decline in MCC relative to CE Loss. 
In contrast, the GAN-based synthetic methods generally enhance accuracy for both majority and minority classes. Nonetheless, the improvements are insufficient for certain minority classes due to the inadequate number of synthetic samples generated for these categories.
As evidenced by the experimental results in Table~\ref{table1}, the re-weighting and re-sampling methods are characterized by lower Macro-F1 and MCC scores, while the GAN-based synthetic methods exhibit reduced B-ACC values. This observation suggests that the predictions generated by these methods tend to yield higher rates of false positives or false negatives. 
Our proposed method, which iteratively incorporates online image synthesis and accuracy adaptive sampling, shows significant improvements for minority classes while maintaining high accuracy for majority classes. Specifically, our approach outperforms competing methods across all evaluation metrics. Specifically, our method achieves a 4.53\% and 3.97\% increase in MCC compared to CE Loss on HAM10000 and APTOS, respectively.

\subsection{Ablation Study}
To evaluate the effectiveness of the proposed online image synthesis (OIS) and accuracy adaptive sampling (AAS), we conducted an ablation study on HAM-10000 and APTOS datasets.
We compared three models, one with offline image synthesis strategy, one with OIS module, and one with both OIS and AAS. The results of these models are presented in the last three rows of Table~\ref{table1}.
For the experiment with the offline setting, we employed the pre-trained diffusion model and classifier to generate images for each class in advance and trained the model with the real training set and the same synthetic images under the same settings. Compared to the results of the baseline method (CE Loss), our model with offline synthesis method obtains improvements of 2.31\%, 3.19\%, and 2.68\% in Macro-F1, B-ACC, and MCC on the HAM10000 dataset, respectively. Moreover, it achieves a little enhancement than Cell-GAN and StyleGAN2-ADA, which indicating the outperformance of the diffusion models over GANs. For the experiments with the OIS strategy only, we sample images per epoch with the same size for each class. Compared to the offline method, the online module provides 0.24\% and 0.22\% enhancement of MCC for HAM10000 and APTOS datasets. 
To further assess the effectiveness of the AAS strategy, we applied it and observed 1.61\% and 0.59\% improvement of MCC for two datasets, respectively.

\subsection{Discussions}
% ----------------------------------------
\begin{figure*}[t]
\centering
\includegraphics[width=.98\textwidth]{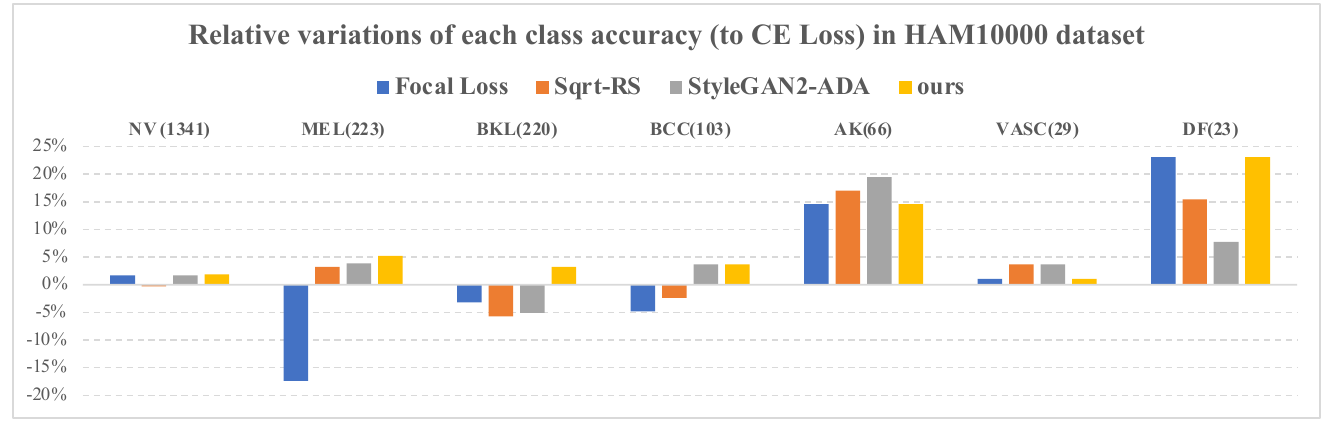}
\caption{
Comparisons of per class accuracy of different methods on HAM10000 dataset. The relative variations of each class accuracy compared to the CE loss are computed. The numbers in brackets following the class name denote the size of samples in that class.
}
\label{fig2}
\end{figure*}
% ----------------------------------------
\begin{figure*}[t]
\centering
\includegraphics[width=.98\textwidth]{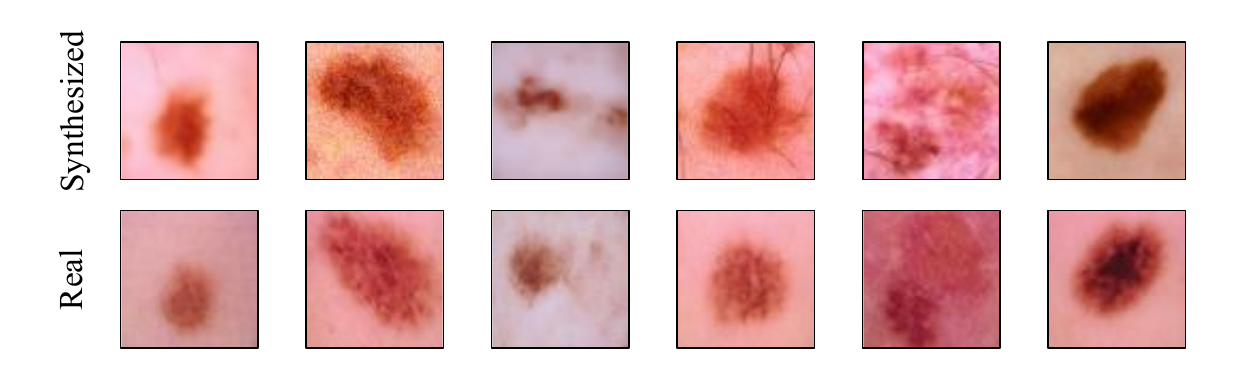}
\caption{
Visualization examples of synthesized images and real images. 
}
\label{fig3}
\end{figure*}
% ----------------------------------------
\subsubsection{Class-wise comparison.}
To evaluate the performance of different methods on each class within the HAM10000 dataset, we computed the relative variations in accuracy for all seven classes in comparison to the CE Loss. The methods considered for comparison are Focal Loss, Sqrt-RS, StyleGAN2-ADA, and our proposed approach. We arranged the seven categories in descending order based on their class sizes, and the results are depicted in Figure~\ref{fig2}. Compared to the CE Loss, Focal Loss exhibits a substantial decline in accuracy for major classes (MEL, BKL, and BCC), despite achieving significant improvements for some minor classes (AK and DR). Furthermore, the Sqrt-RS and StyleGAN2-ADA methods demonstrate suboptimal performance for the BKL and BCC classes. In contrast, our proposed approach consistently exhibits enhancements in accuracy for both major and minor classes when compared to the other methods.

\subsubsection{Visualization of synthetic images.}
To validate the quality of the synthesized images, we present several visual examples of the generated images in Figure~\ref{fig3} and identify the most similar images within the real dataset for each synthetic image. Upon examination, it is evident that the generated samples exhibit diversity while maintaining similarity to the real images. For instance, the first column of images displays a consistent shape, but with varying background illumination levels. The third column of images features a comparable texture, yet with diverse lesion shapes.

\section{Conclusion}
In conclusion, we present a novel approach to tackle the issue of imbalanced classification in medical image analysis. Building on the superior performance of the diffusion model in synthesizing high-quality images, we integrate online image synthesis and accuracy adaptive sampling to iteratively augment training samples that are tailored to the classifier.
Our method demonstrates superior performance when compared to several state-of-the-art approaches on the HAM10000 and APTOS datasets. 
Future research may explore the application of our approach to other medical imaging tasks and the development of more advanced techniques for image synthesis and sampling.

\bibliographystyle{splncs04}
\bibliography{ref}

\begin{thebibliography}{10}
\providecommand{\url}[1]{\texttt{#1}}
\providecommand{\urlprefix}{URL }
\providecommand{\doi}[1]{https://doi.org/#1}

\bibitem{baccouche2020ensemble}
Baccouche, A., Garcia-Zapirain, B., Castillo~Olea, C., Elmaghraby, A.: Ensemble deep learning models for heart disease classification: A case study from mexico. Information  \textbf{11}(4), ~207 (2020)

\bibitem{banik2021mitigating}
Banik, D., Bhattacharjee, D.: Mitigating data imbalance issues in medical image analysis. In: Data preprocessing, active learning, and cost perceptive approaches for resolving data imbalance, pp. 66--89. IGI Global (2021)

\bibitem{carrasco2022assessing}
Carrasco~Limeros, S., Majchrowska, S., Zoubi, M.K., Ros{\'e}n, A., Suvilehto, J., Sj{\"o}blom, L., Kjellberg, M.: Assessing gan-based generative modeling on skin lesions images. In: Machine Intelligence and Digital Interaction Conference. pp. 93--102. Springer Nature Switzerland Cham (2022)

\bibitem{cui2022fmrnet}
Cui, W., Peng, Y., Yuan, G., Cao, W., Cao, Y., Lu, Z., Ni, X., Yan, Z., Zheng, J.: Fmrnet: A fused network of multiple tumoral regions for breast tumor classification with ultrasound images. Medical Physics  \textbf{49}(1),  144--157 (2022)

\bibitem{cui2019class}
Cui, Y., Jia, M., Lin, T.Y., Song, Y., Belongie, S.: Class-balanced loss based on effective number of samples. In: Proceedings of the IEEE/CVF conference on computer vision and pattern recognition. pp. 9268--9277 (2019)

\bibitem{dhariwal2021diffusion}
Dhariwal, P., Nichol, A.: Diffusion models beat gans on image synthesis. Advances in neural information processing systems  \textbf{34},  8780--8794 (2021)

\bibitem{ho2020denoising}
Ho, J., Jain, A., Abbeel, P.: Denoising diffusion probabilistic models. Advances in Neural Information Processing Systems  \textbf{33},  6840--6851 (2020)

\bibitem{kang2019decoupling}
Kang, B., Xie, S., Rohrbach, M., Yan, Z., Gordo, A., Feng, J., Kalantidis, Y.: Decoupling representation and classifier for long-tailed recognition. arXiv preprint arXiv:1910.09217  (2019)

\bibitem{aptos2019-blindness-detection}
Karthik, Maggie, S.D.: Aptos 2019 blindness detection (2019), \url{https://kaggle.com/competitions/aptos2019-blindness-detection}

\bibitem{li2023learning}
Li, J., Cao, H., Wang, J., Liu, F., Dou, Q., Chen, G., Heng, P.A.: Learning robust classifier for imbalanced medical image dataset with noisy labels by minimizing invariant risk. In: International Conference on Medical Image Computing and Computer-Assisted Intervention. pp. 306--316. Springer (2023)

\bibitem{li2022flat}
Li, J., Chen, G., Mao, H., Deng, D., Li, D., Hao, J., Dou, Q., Heng, P.A.: Flat-aware cross-stage distilled framework for imbalanced medical image classification. In: International Conference on Medical Image Computing and Computer-Assisted Intervention. pp. 217--226. Springer (2022)

\bibitem{li2023dynamic}
Li, S., Li, X., Xu, X., Cheng, K.T.: Dynamic subcluster-aware network for few-shot skin disease classification. IEEE Transactions on Neural Networks and Learning Systems  (2023)

\bibitem{lin2017focal}
Lin, T.Y., Goyal, P., Girshick, R., He, K., Doll{\'a}r, P.: Focal loss for dense object detection. In: Proceedings of the IEEE international conference on computer vision. pp. 2980--2988 (2017)

\bibitem{mahajan2018exploring}
Mahajan, D., Girshick, R., Ramanathan, V., He, K., Paluri, M., Li, Y., Bharambe, A., Van Der~Maaten, L.: Exploring the limits of weakly supervised pretraining. In: Proceedings of the European conference on computer vision (ECCV). pp. 181--196 (2018)

\bibitem{rombach2022high}
Rombach, R., Blattmann, A., Lorenz, D., Esser, P., Ommer, B.: High-resolution image synthesis with latent diffusion models. In: Proceedings of the IEEE/CVF conference on computer vision and pattern recognition. pp. 10684--10695 (2022)

\bibitem{saleh2020brain}
Saleh, A., Sukaik, R., Abu-Naser, S.S.: Brain tumor classification using deep learning. In: 2020 International Conference on Assistive and Rehabilitation Technologies (iCareTech). pp. 131--136. IEEE (2020)

\bibitem{shen2023cellgan}
Shen, Z., Cao, M., Wang, S., Zhang, L., Wang, Q.: Cellgan: Conditional cervical cell synthesis for augmenting cytopathological image classification. In: International Conference on Medical Image Computing and Computer-Assisted Intervention. pp. 487--496. Springer (2023)

\bibitem{tan2021equalization}
Tan, J., Lu, X., Zhang, G., Yin, C., Li, Q.: Equalization loss v2: A new gradient balance approach for long-tailed object detection. In: Proceedings of the IEEE/CVF conference on computer vision and pattern recognition. pp. 1685--1694 (2021)

\bibitem{tan2020equalization}
Tan, J., Wang, C., Li, B., Li, Q., Ouyang, W., Yin, C., Yan, J.: Equalization loss for long-tailed object recognition. In: Proceedings of the IEEE/CVF conference on computer vision and pattern recognition. pp. 11662--11671 (2020)

\bibitem{tschandl2018ham10000}
Tschandl, P., Rosendahl, C., Kittler, H.: The ham10000 dataset, a large collection of multi-source dermatoscopic images of common pigmented skin lesions. Scientific data  \textbf{5}(1), ~1--9 (2018)

\bibitem{wang2021detection}
Wang, X., Han, Y., Sun, G., Yang, F., Liu, W., Luo, J., Cao, X., Yin, P., Myers, F.L., Zhou, L.: Detection of the microvascular changes of diabetic retinopathy progression using optical coherence tomography angiography. Translational vision science \& technology  \textbf{10}(7),  31--31 (2021)

\bibitem{yan2022nuclei}
Yan, R., Ren, F., Li, J., Rao, X., Lv, Z., Zheng, C., Zhang, F.: Nuclei-guided network for breast cancer grading in he-stained pathological images. Sensors  \textbf{22}(11), ~4061 (2022)

\bibitem{yang2021ida}
Yang, H., Zhou, Y.: Ida-gan: A novel imbalanced data augmentation gan. In: 2020 25th International Conference on Pattern Recognition (ICPR). pp. 8299--8305. IEEE (2021)

\bibitem{ye2023synthetic}
Ye, J., Ni, H., Jin, P., Huang, S.X., Xue, Y.: Synthetic augmentation with large-scale unconditional pre-training. In: International Conference on Medical Image Computing and Computer-Assisted Intervention. pp. 754--764. Springer (2023)

\bibitem{zhong2023meddiffusion}
Zhong, Y., Cui, S., Wang, J., Wang, X., Yin, Z., Wang, Y., Xiao, H., Huai, M., Wang, T., Ma, F.: Meddiffusion: Boosting health risk prediction via diffusion-based data augmentation. arXiv preprint arXiv:2310.02520  (2023)

\end{thebibliography}

\end{document}